\documentclass[conference]{IEEEtran}
\IEEEoverridecommandlockouts
\usepackage{cite}
\usepackage{amsmath,amssymb,amsfonts}
\usepackage{algorithmic}
\usepackage{graphicx}
\usepackage{textcomp}
\usepackage{xcolor}
\def\BibTeX{{\rm B\kern-.05em{\sc i\kern-.025em b}\kern-.08em
    T\kern-.1667em\lower.7ex\hbox{E}\kern-.125emX}}
\begin{document}

\title{Enhancing UAV Security Through Zero Trust Architecture: An Advanced Deep Learning and Explainable AI Analysis\\
}

\author{\IEEEauthorblockN{Ekramul Haque$^{1}$, Kamrul Hasan$^{1}$, Imtiaz Ahmed$^{2}$, Md. Sahabul Alam$^{3}$, Tariqul Islam $^{4}$  }
\IEEEauthorblockA{$^{1}$ Tennessee State University, Nashville, TN, USA \\
$^{2}$Howard University, Washington, DC, USA \\
$^{3}$California State University, Northridge, CA, USA \\
$^{4}$ Syracuse University, Syracuse, NY, USA \\
Email: $\lbrace$\textit{ehaque1, mhasan1}$\rbrace$@tnstate.edu, \textit{imtiaz.ahmed}@howard.edu,
\textit{md-sahabul.alam}@csun.edu, \textit{mtislam}@syr.edu}}

\maketitle

\begin{abstract}
In the dynamic and ever-changing domain of Unmanned Aerial Vehicles (UAVs), the utmost importance lies in guaranteeing resilient and lucid security measures. This study highlights the necessity of implementing a Zero Trust Architecture (ZTA) to enhance the security of unmanned aerial vehicles (UAVs), hence departing from conventional perimeter defences that may expose vulnerabilities. The Zero Trust Architecture (ZTA) paradigm requires a rigorous and continuous process of authenticating all network entities and communications. The accuracy of our methodology in detecting and identifying unmanned aerial vehicles (UAVs) is 84.59\%. This is achieved by utilizing Radio Frequency (RF) signals within a Deep Learning framework, a unique method. Precise identification is crucial in Zero Trust Architecture (ZTA), as it determines network access. In addition, the use of eXplainable Artificial Intelligence (XAI) tools such as SHapley Additive exPlanations (SHAP) and Local Interpretable Model-agnostic Explanations (LIME) contributes to the improvement of the model's transparency and interpretability. Adherence to Zero Trust Architecture (ZTA) standards guarantees that the classifications of unmanned aerial vehicles (UAVs) are verifiable and comprehensible, enhancing security within the UAV field.

\end{abstract}


\section{Introduction}

Commercial drones, also known as Unmanned Aerial Vehicles (UAVs), are now widely used in several fields, such as delivery services, agriculture, surveillance, and emergency response \cite{1,2,3,4}. This is owing to their versatile applications in the quickly changing environment of unmanned aerial systems.   Nevertheless, the extensive implementation of this technology has brought about intricate security obstacles, especially in areas that require heightened caution.   The significance of tackling these difficulties has been emphasized by recent instances, as seen in the increased attention given to UAV security measures by the U.S. Army and other international institutions\cite{5}. 

The concept of Zero Trust Architecture (ZTA)\cite{6} is a revolutionary way to address the growing security risks posed by unmanned aerial vehicles (UAVs).   ZTA substantially reconceptualizes conventional security models by promoting a constant and thorough verification of all entities inside a network environment, eliminating implicit trust.   This transition is vital within the realm of UAV operations, where the ever-changing and frequently unforeseeable characteristics of flight paths and interactions require a security framework that is more robust and flexible. 

Although there is a clear need for it, the use of ZTA (Zero Trust Architecture) in UAV (Unmanned Aerial Vehicle) security is still in its early stages. One particular problem is establishing continuous authentication mechanisms that can keep up with the changing operational dynamics of UAVs.   Traditional security tactics frequently prove insufficient, either due to their lack of flexibility or failure to effectively handle the full range of possible dangers posed by UAVs. 

The objective of our research is to close this divide by suggesting a sophisticated combination of Deep Learning (DL) and Explainable Artificial Intelligence (XAI) within the ZTA framework for UAV security.   The purpose of this integration is to improve both the precision of UAV identification and the level of transparency and accountability in AI-driven decision-making. This is crucial for establishing confidence in security applications\cite{amin2024empowering}. 

Prior studies have investigated the identification and categorization of UAVs using diverse methodologies.   For instance, Kim et al.\cite{7} and Choi and Oh\cite{8} showcased the utilization of Convolutional Neural Networks (CNNs) with Doppler and micro-doppler images to classify UAVs. Similarly, other researchers have employed deep belief networks and CNNs to analyze surveillance images\cite{9} and audio spectrograms\cite{10}.   However, these researches have mostly concentrated on the detection component without completely incorporating their methods into a broader security framework such as Zero Trust Architecture (ZTA). 

Our technology utilizes Radio Frequency (RF) signals in a DL framework, resulting in an accuracy 84.59\% in differentiating UAV signals from RF noise. Importantly, we broaden the utilization of SHapley Additive exPlanations (SHAP)\cite{11} and Local Interpretable Model-agnostic Explanations (LIME)\cite{12} beyond their typical application. These Explainable Artificial Intelligence (XAI) technologies not only improve the precision of our UAV detection model but also offer a crucial level of interpretability and transparency, allowing stakeholders to comprehend and have confidence in the judgments made by the AI system. 

To summarise, this article greatly enhances the field of UAV security by creatively integrating Zero Trust Architecture (ZTA), Deep Learning (DL), and Explainable Artificial Intelligence (XAI)\cite{amin2024explainable}.  Our methodology tackles the technological and trust-related obstacles in UAV security, facilitating the implementation of stronger and more transparent security protocols in an airspace that is becoming increasingly occupied with drones.

The remainder of this paper is organized as follows: Section
II describes the System Architecture; Section III presents the results from tests conducted by the framework where the proposed architecture was evaluated. Finally, Section IV briefly discusses future enhancements and concluding remarks.

\section{System Architecture}

\subsection{Zero Trust Architecture Framework Description}
In this section, we present the details of our continuous authentication scheme under a zero-trust security framework.

The conventional method of authentication involves validating the legality of an entity at the commencement of a session, hence exposing it to potential security breaches such as hijacking. As suggested in this proposal, continuous authentication involves the ongoing verification of the connecting node's identity during a session. This approach aims to supplement the existing static method rather than replace it entirely.

\begin{figure}[ht]
    \centering
    \includegraphics[height= 5cm, width=\linewidth]{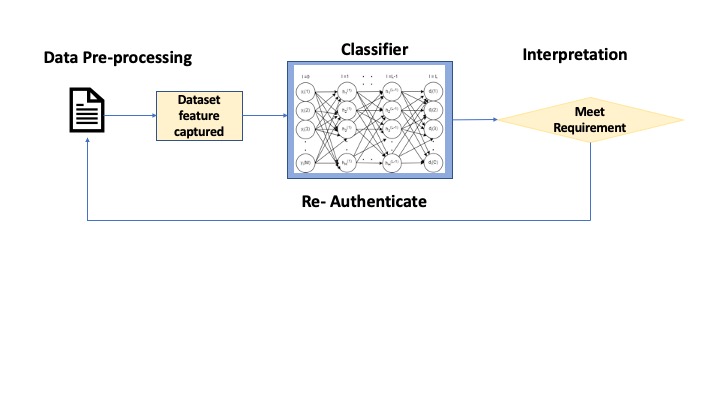}
    \vspace{-2.5cm}
    \caption{Continuous Authentication Workflow}
    \label{fig: ZTA diagram}
\end{figure}

Within this framework, a continuous authentication mechanism is employed to guarantee the security of Unmanned Aerial Vehicles (UAVs). Figure 1 illustrates the workflow of the continuous authentication technique. The procedure above is iteratively conducted at regular intervals of 10 seconds, as determined through empirical observation and analysis. The data features are extracted from the input data. The classifier is trained using the entire dataset. A safe ZTA framework for UAVs starts with determining the type of UAV in use. Our DNN model classifies UAVs by RF signals, distinguishing brands and types. This classification requires a baseline understanding of the UAV's properties and predicted behavior. Classification is a necessary but basic stage in authentication. Correctly recognizing the UAV type allows the system to apply relevant security policies and permissions.

Now, we will delve into our DNN classifier architecture. Figure 2 presents the architecture, an AI-driven framework for UAV classification. It integrates PCA, drones, LIME, and SHAP to enhance the DL model's credibility using explainable ML tools.

The process begins by preparing RF data, converting it from time to frequency domain using DFT. The frequency components act as inputs, with power spectra as critical features, forming the basis for DNN training.

\begin{figure}[htbp]
    \centering
    \includegraphics[height = 5cm,width= 0.98\linewidth]{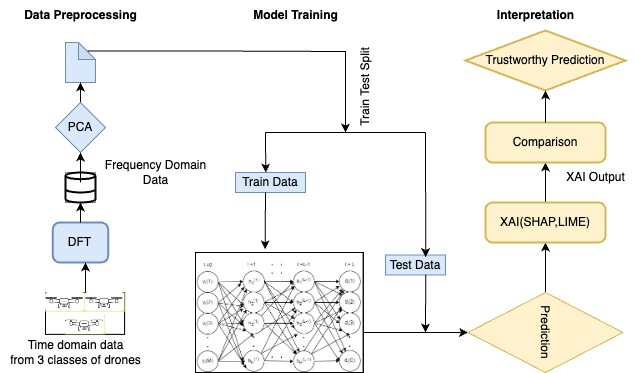}
    \caption{Illustration of the Interpretable AI Framework for
Advanced Classification of UAVs.}
    \label{fig: AI diagram}
\end{figure}

To address dataset complexity, PCA reduces dimensionality, streamlining computational needs and removing redundancy. The refined dataset then trains the DNN model to classify drones into four categories.

For transparency, we employ XAI techniques, specifically SHAP and LIME, ensuring transparent decision-making and boosting user trust in model classifications.

\subsection{Analysis of the Data set}

We used the vastly popular DroneRF dataset\cite{13}, which contains captured RF signals from drones of three kinds. Segmenting and storing the DroneRF dataset in CSV format prevents memory overflow and ensures program compatibility. The model divides drones into four categories based on data from three brands and the lack of drones.

The main problem was translating time-domain data to frequency domain for better analysis.
Achieved this conversion using the DFT. Our deep learning model relies on power spectra, particularly frequency components, as its fundamental properties.

\subsection{Data Preprocessing}

Each archived RF segment signal is transformed using the MATLAB Fast Fourier Transform (FFT) function with 2048 frequency divisions (M = 2048). The frequency-domain RF signals are represented by equations \ref{eq:(1)} \& \ref{eq:(2)} as\cite{1}:
\begin{equation}
\label{eq:(1)}
y_i^{(L)}(m) = \Bigg\lvert\sum_{n=1}^{N} x_i^{(L)}(n) \exp \left(\frac{-j2\pi m(n-1)}{N}\right)\bigg\rvert
\end{equation}

\begin{equation}
\label{eq:(2)}
y_i^{(H)}(m) = \Bigg\lvert\sum_{n=1}^{N} x_i^{(H)}(n) \exp \left(\frac{-j2\pi m(n-1)}{N}\right)\bigg\rvert
\end{equation}

where $x_i^{(L)}$ denotes the RF segment $i$, coming from the first RF receiver that captures the lower half of the RF spectrum, $x_i^{(H)}$ is the $i$th RF segment coming from the second RF receiver that captures the upper half of the RF spectrum, $y_i^{(L)}$ and $y_i^{(H)}$ are the spectra of the $i$th segments coming from the first and second RF receivers respectively, $n$ and $m$ are the time and frequency domain indices, $N$ is the total number of time samples in the RF segment $i$, and $|$ is the magnitude operator used to compute the power spectrum. Note that $y_i^{(L)}$ and $y_i^{(H)}$ contain only the positive spectra of $x_i^{(L)}$ and $x_i^{(H)}$ to guarantee non-redundant and concise spectral projections. Then, we combine the transformed signals of both receivers to create the entire RF spectrum by equations \ref{eq:(3)} \& \ref{eq:(4)}, as follows:
\begin{equation}
\label{eq:(3)}
    y_i = [y_i^{(L)},cy_i^{(H)}]
\end{equation}
and
\begin{equation}
\label{eq:(4)}
    c = \frac{\sum_{q=0}^{Q} y_i^{(L)}(M-q)}{\sum_{q=0}^{Q} y_i^{(H)}(q)}
\end{equation}
where $c$ is a normalization factor calculated as the ratio between the last $Q$ samples of the lower spectra, $y_i^{(L)}$, and the first $Q$ samples of the upper spectra, $y_i^{(H)}$, and $M$ is the total number of frequency bins in $y_i$. The normalization factor, $c$, ensures spectral continuity between the two halves of the RF spectrum as they were captured using different devices; hence, a spectral bias is inevitable. Note that $Q$ must be relatively small to successfully stitch the two spectra and be large enough to average out any
random fluctuations, e.g. $Q$ = 10 for $M$ = 2048.

\subsection{Model Architecture}
The proposed model can detect drones and differentiate among the RF spectra of distinct drones. A DNN comprises an input layer, different hidden layers, and an output layer.

Using the following expressions\cite{14}, one can articulate the input-output relationship of a DNN by equations \ref{eq:(5)} \& \ref{eq:(6)}: 
\begin{equation}
\label{eq:(5)}
z_i^{(l)} = f^{(l)}([W^{l}z_i^{(l-1)} + b^{(l)})
\end{equation}

\begin{equation}
\label{eq:(6)}
W^{(l)} = 
\left[
\begin{aligned}
w_{11}^{(l)} &\hspace{2em} w_{12}^{(l)} \hspace{2em} \ldots \hspace{2em} w_{1H^{(l-1)}}^{(l)} \\
w_{21}^{(l)} &\hspace{2em} w_{22}^{(l)} \hspace{2em} \ldots \hspace{2em} w_{2H^{(l-1)}}^{(l)} \\
&\vdots \hspace{3em} \vdots \\
w_{H^{(l)}1}^{(l)} &\hspace{2em} w_{H^{(l)}2}^{(l)} \hspace{2em} \ldots \hspace{1em} w_{H^{(l)}H^{(l-1)}}^{(l)}
\end{aligned}
\right]
\end{equation}

where $z^{(l-1)}$ i is the output of layer $l$-1 and the input to layer $l$; $z_i^{(l)}$ is the output of layer $l$ and the input to layer $l$ + 1; $z_i^{(0)}$ = $y_i$ is the spectrum of the RF segment $i$; $z_i^{(L)}$ = $d_i$ is the classification
vector for the RF segment $i$; $W^{(l)}$ is the weight matrix of layer $l$;$w_pq^{(l)}$ is the weight between the $p$th neuron of layer $l$ and the $q$th neuron of layer $l$ - 1; $b^{(l)}$ = $[b_1^{(l)}$, $b_2^{(l)}$, ....$b_{H^{(l)}}^{(l)}]^T$
is the bias vector of layer $l$; f ($l$) is the activation function of layer $l$; $l$ = 1, 2, . . . , $L$; $L$ - 1 is the total number of hidden layers; $H^{(l)}$ is the total number of neurons in layer $l$; $H^{(0)} = M$; $H^{(L)} = C$; and $C$ is the number of classes in the classification vector, $d_i$\cite{14}. Note that $f$ can be any linear or nonlinear function; however, the rectified linear unit (ReLU) and the softmax functions, expressed in \ref{eq:(7)} and \ref{eq:(8)} respectively, are typical choices that have demonstrated promising results.

\begin{equation}
\label{eq:(7)}
f(x) = 
\begin{cases}
x & \text{if } x > 0 \\
0 & \text{if } x \leq 0
\end{cases}
\end{equation}

\begin{equation}
\label{eq:(8)}
f(x) = \frac{1}{1+e^{-x}}
\end{equation}

The DNN's weights and biases are determined by a supervised learning method that minimizes classification error. The minimization is carried out by a gradient descent algorithm, which calculates the gradient by backpropagation. The classification error of the system is represented by the categorical cross-entropy error in equation \ref{eq:(9)}:

\begin{equation}
\label{eq:(9)}
    L = -\frac{1}{N} \sum_{i} \sum_{j} y_{ij} \log(p_{ij})
\end{equation}

where $N$ is the total number of samples, $y_ij$ is the true label for class $j$ of the ith sample, often one-hot encoded, $p_ij$ is the predicted probability for class $j$ of the ith sample, $\sum_{i}$ denotes the sum over samples, $\sum_{j}$ denotes the sum of classes.

In this study, a DNN is trained and evaluated using the developed RF database to detect the presence of a drone and determine its class.

\subsection{Generating Explanations}
SHAP calculates the contribution of each feature to the model's prediction. The consequence of every potential combination of characteristics is analyzed to determine the feature relevance. If $n$ features exist, SHAP creates $2n$ unique prediction models. The dataset is the same in all models; the only difference is the number of features evaluated. The difference between these models' predictions will aid in calculating the overall relevance of the characteristic. Here's how SHAP values are calculated. For an arbitrary model and instance, the SHAP value $\phi$ for a feature $i$ is given by equation \ref{eq:(10)} like \cite{11}:

\begin{align}
\label{eq:(10)}
\phi_i = \sum_{S \subseteq F \setminus \{i\}} & \frac{|S|!(|F| - |S| - 1)!}{|F|!} \nonumber \\ 
& \times \left[f_{S \cup \{i\}}(x_{S \cup \{i\}}) - f_S(x_S)\right]
\end{align}

where $f$ is the prediction model, $F$ is the set of all features, and $S$ is a subset of $F$, not containing feature $i$. $U$ represents the union of sets. $\lvert S \rvert$ is the size of set $S$. The sum over all subsets $S$ of $F$ does not contain feature $i$.

LIME provides localized interpretations of classifier predictions. Starting with an instance, $x$, it generates a perturbed dataset around $x$, predicts probabilities for each instance using the original model, and assigns weights based on proximity to $x$. An interpretable model, like linear regression, is then fitted to this dataset, approximating the original model's behavior near $x$.

Given $g$ as the interpretable model, $f$ as the complex model, and $\Omega(g)$ as a complexity measure, LIME optimizes:
\begin{equation}
\label{eq:(11)}
\sum_{i} \pi_x(z_i) (f(z_i) - g(z_i))^2 + \Omega(g)
\end{equation}
where $\pi_x(z_i)$ measures proximity between $x$ and samples $z$. The objective is a simple model with minimized prediction loss.

\section{Experimental Analysis}

\subsection{Principal Component Analysis}
The tests preserved 95\% of the variation in the baseline dataset. We chose this solution to strike a balance between processing efficiency and data integrity. Initial trials indicated that retaining variance was an acceptable trade-off, preserving important training properties while reducing processing needs.

\begin{table}[ht]
\centering
\caption{Comparison between computational time}
\label{tab: comparison}
\resizebox{0.9\linewidth}{!}{
\begin{tabular}{|c|c|c|}
  \hline
   & 
   \begin{tabular}{@{}c@{}}Features \end{tabular}
   & 
   \begin{tabular}{@{}c@{}}Computational Time\\(ms)\end{tabular} \\
  \hline
  Without PCA\cite{1} & 2047 & 77.2 \\
  \hline
  With PCA & 688 & 51.6 \\
  \hline
\end{tabular}}
\end{table}

From Table \ref{tab: comparison}, we can see that using PCA, we reduced the number of features to 688 while maintaining 95 \% of the original data's variance, which made the algorithm significantly faster.

\subsection{Experimental Result}

In this experiment, the DNN is trained by an Adam optimizer to minimize the categorical cross-entropy of classification using the following parameters: 3 hidden fully-connected layers ($L-1 = 3)$, 256, 128 and 64 total number of neurons at the first, second and third hidden layers respectively $H(1)$ = 256, $H(2)$ = 128, $H(3)$ = 64), 200 total epochs, 10 batch size, $f$ is the ReLU function for the hidden layers and the softmax function for the output layer. Each network's classification performance is validated utilizing a stratified 10-fold cross-validation procedure (K = 10). It is to be mentioned here that, in their work on UAV detection\cite{15}, M.F. Al-Sa'd et al. used mse for loss and sigmoid function for the output layer. However, we used categorical cross-entropy as a loss function and softmax function for the output layer. As a result, even after using PCA, we got better accuracy. Table \ref{tab: comprehensive} shows the complete evaluation results' comparison for the DNN models based on accuracy, precision, recall, and F1 score.

\begin{table}[ht]
\caption{Comprehensive Comparison of Models Performance}
\label{tab: comprehensive}
\resizebox{0.9\linewidth}{!}{
\begin{tabular}{|c|c|c|c|c|}
  \hline
  \begin{tabular}{@{}c@{}}\end{tabular} 
  &
  \begin{tabular}{@{}c@{}}Accuracy\\(\%)\end{tabular} & 
  \begin{tabular}{@{}c@{}}Precision\\(\%)\end{tabular} & 
  \begin{tabular}{@{}c@{}}Recall\\(\%)\end{tabular} 
  &
  \begin{tabular}{@{}c@{}}F1\\(\%)\end{tabular} \\
  \hline
  Without PCA\cite{15} & 84.50 & 92.02 & 76.96 & 79.45 \\
  \hline
  With PCA & 84.59 & 90.32 & 78.39 & 81.02 \\
  \hline
\end{tabular}}
\end{table}

The table reveals that we have improved in every metric except for precision. 

\subsection{Deep Neural Network Performance: Confusion Matrix Analysis}

\begin{figure}[htbp]
  \centering
  \includegraphics[width=\linewidth]{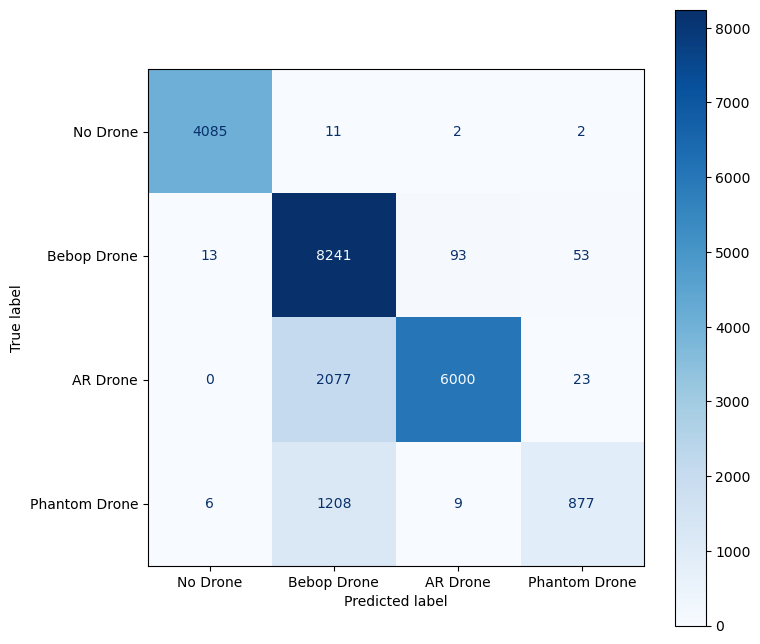}
  \caption{Confusion Matrix for Drone Classification}
  \label{fig: confusion}
\end{figure}

Figure \ref{fig: confusion} represents a confusion matrix that shows the performance of the DNN that classifies the drones through the test dataset. The matrix demonstrates the model's capability to predict the class of drones accurately. The model correctly predicted No Drone cases 4085 times, Bebop Drones 8241 times, AR Drones 6000 times, and Phantom Drones 877 times. 

\subsection{Interpreting Predictions with 
SHAP }

\begin{figure}[htbp]
  \centering
   \includegraphics[width=\linewidth]{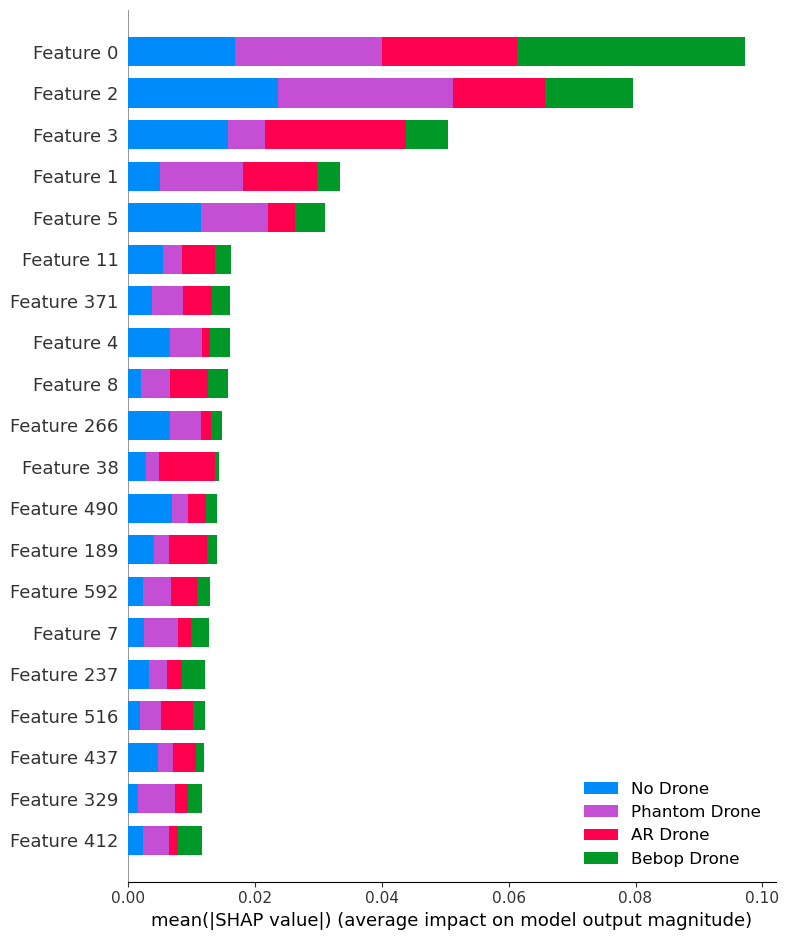}
  \caption{SHAP summary plot}
  \label{fig: summary plot2}
\end{figure}

The SHAP summary plot is a global explanation of a model that incorporates the importance and effect of features. The features are on the $Y$ axis, and the $X$ axis represents the values. The use of colours indicates different drones. A summary plot shows random features and their contributions toward 4 classes as bars. Figure \ref{fig: summary plot2} shows that feature 0 contributes most to Bebop Drone. Similarly, feature 2 has the most influence over the Phantom Drone. 

\subsection{Interpreting Predictions with LIME}
\begin{figure}[htbp]
  \centering
  \includegraphics[width=\linewidth]{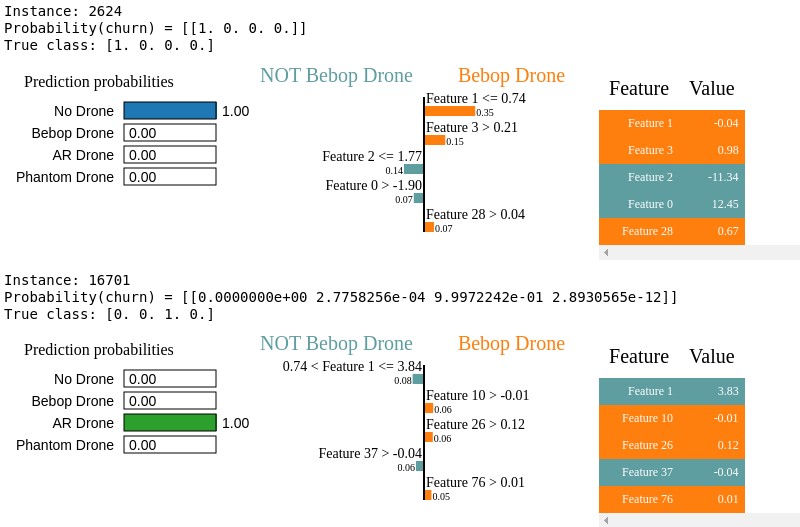}
  \caption{Explaining individual prediction of deep learning classifier using LIME.}
  \label{fig: LIME}
\end{figure}

LIME produces regional explanations. Figure \ref{fig: LIME} illustrates how to determine whether a classification result is No Drone, Bebop, AR' or Phantom, along with the probability and original instance values. Here, we showed two instances. The instances are shown Bebop and NOT Bebop because the classifier finds it most crucial to decide whether the instances are Bebop or not. Colors are used to indicate the class a character belongs to. While orange attributes contribute to the Bebop category, blue features contribute to the Not Bebop category.

We can see distinct characteristics justify different classifications. In instance 2624, Features 2 and 3 contribute to the prediction that there is no drone, whereas in instance 19701, features 1 and 37 contribute to the prediction that there is an AR Drone.

\section*{Conclusion}

The training dataset limits our current model's ability to classify UAVs based on RF signals. The dataset only includes RF signals from three drone kinds, limiting the model's capacity to generalize to new drone types. When encountering a drone not in the training dataset, our machine may misclassify, assign it to a known category, or lose confidence in its predictions.
This limitation is significant since real-world deployment of such a system will face more UAV kinds. We must overcome this constraint to improve the model's utility and resilience.
We recommend adding an anomaly detection algorithm to our future work to overcome this constraint and make our model more applicable in varied operational situations. First, this algorithm would identify when the model encounters a drone type that significantly deviates from the known categories, and second, it would flag these instances for manual review or classify them as 'unknown'. Such an approach would improve the model's accuracy in known scenarios and allow it to adapt to new conditions, increasing its real-world applicability.
Future studies will expand the dataset to include more drone kinds. The model will learn more RF signal characteristics with this increase, enhancing classification accuracy and generalizability. An anomaly detection component and a richer, more varied dataset will improve our Zero Trust Architecture UAV security model."

\bibliographystyle{ieeetr}
\bibliography{references.bib}
\end{document}